# End-to-End Learning for the Deep Multivariate Probit Model

Di Chen [1]    Yexiang Xue [1]    Carla Gomes [1]

## Abstract

The multivariate probit model (MVP) is a popular classic model for studying binary responses of multiple entities. Nevertheless, the computational challenge of learning the MVP model, given that its likelihood involves integrating over a multidimensional constrained space of latent variables, significantly limits its application in practice. We propose a flexible deep generalization of the classic MVP, the Deep Multivariate Probit Model (DMVP), which is an end-to-end learning scheme that uses an efficient parallel sampling process of the multivariate probit model to exploit GPU-boosted deep neural networks. We present both theoretical and empirical analysis of the convergence behavior of DMVP's sampling process with respect to the resolution of the correlation structure. We provide convergence guarantees for DMVP and our empirical analysis demonstrates the advantages of DMVP's sampling compared with standard MCMC-based methods. We also show that when applied to multi-entity modelling problems, which are natural DMVP applications, DMVP trains faster than classical MVP, by at least an order of magnitude, captures rich correlations among entities, and further improves the joint likelihood of entities compared with several competitive models.

## 1. Introduction

Understanding multi-entity interactions is a central question in many real-world applications. For example, in computational sustainability (Gomes, 2009; MacKenzie et al., 2004), it is important to understand the spatial distribution of species and how species interact with each other and their environment, for developing conservation plans. In computer vision, the detections of multiple objects are often correlated because of the shared background and scenario (Wang et al., 2016). In natural language processing, a text often has several correlated labels in terms of its topic, emotion, and semantic meaning (Nam et al., 2014). The multivariate probit model (MVP) (Ashford & Sowden, 1970) is a popular classic model for studying interactions of multiple entities. Nevertheless, learning the multivariate probit model is challenging because it involves the integration of a multivariate normal distribution over a constrained space.

A classic approach for optimizing the MVP model is Bayesian Inference (Chib & Greenberg, 1998; Tabet, 2007), where the posterior distribution is simulated by Markov Chain Monte Carlo (MCMC) methods (Jeliazkov & Hee Lee, 2010) and the maximum likelihood estimates are obtained by a Monte Carlo version of the Expectation Maximization (EM) algorithm. These approaches require the simulation of observations from a multivariate truncated normal distribution involving an arbitrary covariance matrix. Although observations from a multivariate truncated normal distribution can be sampled from a sequence of univariate truncated normal distributions (Genz, 1992), the computational effort is rather heavy for high dimensional problems. Extensions of the classic MVP in specific domains have been proposed under specific assumptions of the covariance matrix (see e.g., (Song & Lee, 2005; Young et al., 2009)). Recent approaches for computing the maximum likelihood of MVP have been proposed using the first-order gradients and the second-order information (Chen et al., 2017; Mandt et al., 2017). Those approaches integrate MCMC methods and the numerical estimation of the multivariate probit (Genz, 1992), which is based on an importance sampling using the truncated normal distribution.

The accessibility of massive contextual data, as well as the success of deep learning, provide additional opportunities and challenges for boosting MVP. On the one hand, massive contextual data, such as millions of high-resolution images, create the possibility of improving predictive performance, particularly when integrated with deep neural networks, which are remarkably powerful for extracting high-level features from raw data. On the other hand, a scalable learning scheme, which integrates well with parallelized infrastructure such as GPUs, is needed to take advantage of various deep neural networks as well as the massive contextual data. Unfortunately, the classical approaches such as Bayesian inference or previous gradient-based methods, inevitably contain sequential inferences, such as MCMC simulations, which is typically not easy to implement on GPUs. A recent approach called Multi-Entity Dependency Learning via Conditional Variational Auto-encoder (MEDL_CVAE) (Tang et al., 2017) is compatible with deep neural networks and exploits GPUs, with competitive wall-clock training time, but suffers from two key limitations. On one hand, MEDL_CVAE learns the joint likelihood by optimizing the variational lower bound of the joint likelihood but has no guarantee concerning the gap between the lower bound and the true likelihood. A second limitation is that the empirical optimization of the variational lower bound of MEDL_CVAE suffers from the KL-vanishment problem, which is a known problem in applications based on variational auto-encoder. As a result, when integrating it with powerful deep neural networks such as Convolutional Neural Networks, the KL-term decreases dramatically to zero, which causes serious overfitting problems that restrict its performance.

---

[1]Computer Science Department, Cornell University, Ithaca, NY, US 14850. Correspondence to: Di Chen <di@cs.cornell.edu>.

We propose **a novel end-to-end learning scheme for the Deep Multivariate Probit Model (DMVP), which is scalable and flexible with various deep neural networks.** Specifically: **(1)** We introduce the *Deep Multivariate Probit Model* (DMVP), a deep generalization of classic MVP, in which a flexible deep neural network is embedded to extract the high-level features from the raw data. **(2)** We propose an efficient parallel sampling process, which transforms the integration over a high-dimensional constrained space into an expectation over the residual multivariate normal distribution with a variance strictly lower than the rejection sampling, tightly integrates with various deep neural networks, and can be implemented end-to-end on GPUs. **(3)** We provide both a theoretical and an empirical analysis of the convergence behavior of the sampling process embedded in DMVP. We provide theoretical convergence guarantees for DMVP as well as a numerical analysis of the convergence behavior based on a tighter bound, which is much closer to the empirical results. Our theoretical bound also sheds light on the trade-offs between performance and convergence. **(4)** We apply DMVP to three multi-entity modelling problems. In the first application, we use the crowdsourced *eBird* dataset combined with the National Land Cover Database for the U.S (NLCD) (Homer et al., 2015) and satellite images to study the interaction among multiple species. In the second application, we study the deforestation and human encroachment in Amazon rainforest with high-resolution satellite images. In the last application, we study the associated concepts of real-world web images using the NUS-WIDE-LITE web image dataset collected from Flickr (Chua et al., July 8-10, 2009).
**Preview of results:** We show that our DMVP (a) trains significantly faster than classic MVP models using the end-to-end learning scheme fully implemented on GPUs; (b) captures correlations among multiple entities in all applications; and (c) outperforms the approaches that assume the independence among entities conditioned on contextual data, classic MVP models, recent gradient-based MVP methods, and the recent variational approach MEDL_CVAE.

## 2. Preliminaries

### 2.1. Notations

We use $\phi(x; \mu, \Sigma)$ and $\Phi(x; \mu, \Sigma)$ to denote the density function and the cumulative distribution function of the multivariate normal distribution with mean $\mu \in \mathbb{R}^l$ and covariance $\Sigma \in \mathbb{R}^{l \times l}$, i.e,

$$\phi(x; \mu, \Sigma) = \frac{1}{(2\pi)^{l/2}|\Sigma|^{1/2}} e^{-\frac{1}{2}(x-\mu)^T \Sigma^{-1}(x-\mu)} \quad (1)$$

$$\Phi(x; \mu, \Sigma) = \int_{-\infty}^{x_1} ... \int_{-\infty}^{x_l} \phi(s; \mu, \Sigma).\mathrm{d}s_1...\mathrm{d}s_l \quad (2)$$

where $|\cdot|$ denotes the determinant of a matrix.

For the sake of simplicity, we use $\Phi(x)$ to denote the CDF of one-dimensional standard normal distribution.

For the comparison between vectors, we use "$\preccurlyeq$" to denote the "element-wise less or equal to", i.e,

$$a \preccurlyeq b \text{ iff } \forall i, a_i \leq b_i \quad (3)$$

### 2.2. Deep Multivariate Probit Model

The multivariate probit model (MVP) is described in terms of a multivariate normal distribution of the underlying latent variables that are manifested as binary responses through a threshold specification. More specifically, given the dataset $D = \{(x_i, y_i)|i = 1, ..., N\}$, where $x_i \in \mathbb{R}^m$ is the $m$-dimensional contextual data and $y_i \in \{0, 1\}^l$ is the $l$-dimensional binary label, MVP maps the Bernoulli distribution of each binary label $y_{i,j}$ to a sequence of latent variables $r_i = (r_{i,1}, ..., r_{i,l})$ through the threshold 0, where $r_i$ is subject to a multivariate normal distribution, i.e,

$$\begin{aligned} \Pr(y_{i,j} = 1|x_i) &= \Pr(r_{i,j} > 0) \\ \Pr(y_{i,j} = 0|x_i) &= \Pr(r_{i,j} \leq 0) \\ \text{where } r_i &\sim N(\mu(x_i), \Sigma). \end{aligned} \quad (4)$$

More specifically, the marginal likelihood is,

$$\Pr(y_{i,j} = 1|x_i) = \Phi\left(\frac{\mu(x_i)_j}{\sqrt{\Sigma_{j,j}}}\right)$$

$$\Pr(y_{i,j} = 0|x_i) = \Phi\left(-\frac{\mu(x_i)_j}{\sqrt{\Sigma_{j,j}}}\right),$$

and the joint likelihood is,

$$\Pr(y_i|x_i) = \int_{A_1} ... \int_{A_l} \phi(s; \mu(x_i), \Sigma).\mathrm{d}s_1...\mathrm{d}s_l$$

Here $A_j = \begin{cases} (-\infty, 0] \text{ if } y_{i,j} = 0 \\ [0, +\infty) \text{ if } y_{i,j} = 1 \end{cases} \quad (5)$

Let $D^i = \text{diag}(2y_i - 1) \in \{-1, 0, 1\}^{l \times l}$ which is a diagonal matrix using vector $2y_i - 1$ as its diagonal. Then, we can further reduce formula (5) into the CDF of a multivariate normal distribution, using the affine transformation, i.e.,

$$\Pr(y_i | x_i) = \Phi(0; -\mu_i', \Sigma_i'), \tag{6}$$
$$\text{where } \mu_i' = D^i \mu(x_i) \text{ and } \Sigma_i' = D^i \Sigma D^i.$$

Learning the classic MVP model involves estimating the coefficient $W$ of the linear function $\mu(x_i) = Wx_i$ and the covariance matrix $\Sigma$. Usually, both the coefficient matrix $W$ and the covariance matrix $\Sigma$ are learnt from data, but in some cases the variance matrix $\Sigma$ can be computed directly from data. For example, the model in (Mandt et al., 2017) used linear kernel for the covariance matrix, where the covariance matrix is the sum of a linear kernel matrix and a diagonal noise matrix computed from the raw input data.

Taking advantage of the successful development of deep learning, we introduce the Deep Multivariate Probit Model (DMVP), which is a deep generalization of the classic MVP, where $\mu(x_i)$ changes from the linear function $Wx_i$ to a non-linear function $\theta(x_i)$, learnt via a deep neural network, and the covariance matrix $\Sigma$ is always learnt from the data. In this way, the DMVP obtains the flexibility as well as the predictive power of various deep neural networks while modelling the correlations of multiple entities by fitting the correlations of the latent variables.

## 3. End-to-End Learning for DMVP

The generic learning methods of MVP are maximum-a-posteriori estimation and maximum likelihood estimation. Because we have introduced the deep neural networks into the DMVP, we train DMVP by maximizing the log-likelihood, which is the most commonly used method for the training of neural networks, i.e.,

$$\underset{\theta, \Sigma}{\text{argmax}} \sum_i \log \Pr(y_i | x_i) = \underset{\theta, \Sigma}{\text{argmax}} \sum_i \log \Phi(0; -\mu_i', \Sigma_i').$$

The difficulties with respect to learning the DMVP are mainly due to the computation of equation (6) as well as its gradients, which are obtained by integrating over a high-dimensional constrained space of latent variables. As pointed out by (Magid, 1994), there is no closed form solution for equation (6), and to date can only be estimated via sampling methods.

### 3.1. End-to-End Sampling Process for DMVP

The vanilla rejection sampling estimates $\Phi(0; -\mu_i', \Sigma_i')$ by counting the rate that a sample $r$ from $N(-\mu_i', \Sigma_i')$ satisfies $r \preccurlyeq 0$. However, because the value of $\Phi(0; -\mu_i', \Sigma_i')$ could be exponentially small, on average, it could take exponentially many trials to get merely one trial that satisfies the condition. One straightforward solution for this estimation, which has been adopted in (Chen et al., 2017), is to use the MCMC approaches to estimate the distribution over the truncated high-dimensional space. Another importance sampling method proposed by (Genz, 1992) uses Cholesky factorization to compute the equation (6). This method transforms the sampling of a truncated multivariate normal distribution into the sampling of a sequence of univariate truncated normal distributions, where the truncation of each univariate normal distribution depends on the samples of all preceding random variables. Because both the MCMC method and the importance sampling require a sequentially dependent sampling, they cannot easily integrate with parallelized deep learning infrastructure such as GPUs. Therefore, we propose a novel parallel sampling method to address this approximation problem.

Though there is no closed form for computing the CDF of a general multivariate normal distribution, the one-dimensional CDF $\Phi(x)$ has very accurate estimation (Cody, 1969), which has been implemented in almost all machine learning tools. Inspired by this fact, we decompose the covariance matrix $\Sigma$ into $V + \Sigma_r$, where $V$ is a diagonal positive definite matrix and $\Sigma_r$ is the residual covariance matrix, so that a random variable $r \sim N(0, \Sigma)$ can be decomposed as the subtraction of two random variable $z \sim N(0, V)$ and $w \sim N(0, \Sigma_r)$, i.e., $r = z - w$. Thus, the estimation of $\Phi(0; -\mu, \Sigma)$ in equation (6) can be transformed into the expectation of the product of $l$ one-dimensional CDF's, conditioned on the residual multivariate normal distribution $w$, i.e.,

$$\Phi(0; -\mu, \Sigma) = \Pr(r - \mu \preccurlyeq 0) \quad r \sim N(0, \Sigma)$$
$$= \Pr(z - w \preccurlyeq \mu) \quad z \sim N(0, V), \ w \sim N(0, \Sigma_r)$$
$$= E_{w \sim N(0, \Sigma_r)}[\Pr(z \preccurlyeq (w + \mu) | w)] \quad z \sim N(0, V)$$
$$= E_{w \sim N(\mu, \Sigma_r)} \left[ \prod_{j=1}^{l} \Phi \left( \frac{w_j}{\sqrt{V_{j,j}}} \right) \right]$$
$$= E_{w \sim N(V^{-1/2}\mu, \ V^{-1/2}\Sigma_r V^{-1/2})} \left[ \prod_{j=1}^{l} \Phi(w_j) \right]$$
$$\approx \frac{1}{M} \sum_{k=1}^{M} \left( \prod_{j=1}^{l} \Phi \left( w_j^{(k)} \right) \right). \tag{7}$$

Knowing that the role of the parameter $V$ is to rescale the sample $w_j$, without loss of generality, we can assume that $V$ is an identity matrix and directly learn the "rescaled" residual multivariate normal distribution. That is, **in the rest of the paper as well as the Figure (1), we use the identity matrix $I$ to replace $V$.**

**Main idea:** The high-level idea of our end-to-end learning scheme for DMVP is based on the transformation shown in equation (7). The intuition behind our transformation is similar to the *Rao-Blackwell theorem* (Blackwell, 1947), which improves an estimator by computing its expectation, conditioned on a sufficient statistic. In our case, instead of using a sufficient statistic, we use the residual distribution $w$, which fully captures the correlations of the original multivariate distribution. Conceptually, given input features $x_i$ and labels $y_i$, DMVP first learns the mean and the covariance of the residual multivariate normal distribution via a deep neural network. Then, DMVP samples batches of independent samples $w^{(k)}$ from the residual multivariate normal distribution and uses equation (7) to compute the estimation of the joint likelihood. **This sampling process outperforms previous estimation methods in several aspects. First**, the process samples from an explicit distribution, which is significantly more efficient than MCMC-based methods, which need to burn a lot of intermediate samples to reach the implicit distribution. (We show the experimental results in the section 5.) **Second**, the variance of this sampling process is strictly smaller than the vanilla rejection sampling (See appendix), therefore DMVP requires fewer samples, since every sample from this process provides non-trivial information. **Third**, this sampling process can be implemented in parallel on GPUs, which is not the case for MCMC.

Figure (1) depicts the detailed learning framework implemented in DMVP. The feature network, composed of multi-layer convolutional networks or fully connected networks, extracts high-level features from the contextual data source to learn the $\mu$ for each data point. The choice of the feature network depends on the type of contextual data and the problem, but is flexible enough to be any structure that could be boosted by GPUs. In DMVP, the residual covariance matrix $\Sigma_r$ is a global parameter, which is learned from random initialization and shared by all data points. To ensure that $\Sigma_r$ is a semi-positive definite matrix, we actually form the residual covariance matrix by the product of one matrix and its transpose, i.e., $\Sigma_r = \Sigma_r^{1/2}(\Sigma_r^{1/2})^T$. The random variable generator generates batches of the standard normal distributed random variable $z^{(k)}$ in parallel on GPUs. Then, using $\Sigma_r^{1/2}$, $\mu$, and the diagonal matrix $D^i$ corresponding to $y_i$, DMVP computes batches of samples $w^{(k)} = D^i(\mu + \Sigma_r^{1/2} z^{(k)})$. According to the affine transformation of the normal distribution, the samples $\{w^{(k)}\}$ are subject to the multivariate normal distribution $N(D^i\mu, D^i\Sigma_r D^i)$, which is the desired residual multivariate normal distribution as derived in equation (6) and (7). Because there is no dependency among those samples, all the operations described above could be computed in parallel using tensor operations. Therefore, we can integrate DMVP with various deep neural networks and implement it end-to-end on GPUs using popular machine learning packages (such as Tensorflow or Pytouch).

### 3.2. Theoretical Analysis of the DMVP's Convergence Behavior

In terms of the convergence behavior of this sampling process, we provide a theoretical analysis with respect to the estimation error. Since the estimate $\prod_{j=1}^{l} \Phi\left(w_j^{(k)}\right)$ is bounded between 0 and 1, Hoeffding's inequality guarantees exponentially fast convergence in $M$ between the r.h.s of equation (7) and $\Pr(y_i|x_i)$, i.e.,

$$\Pr\left[\left|\frac{1}{M}\sum_{k=1}^{M}\prod_{j=1}^{n}\Phi(w_{i,j}^{(k)}) - \Pr(y_i|x_i)\right| \geq \epsilon \Pr(y_i|x_i)\right]$$
$$\leq 2e^{-M\epsilon^2 \Pr^2(y_i|x_i)}. \tag{8}$$

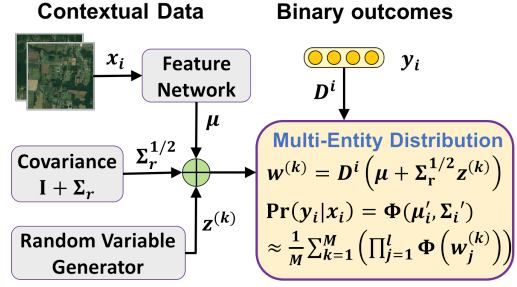

*Figure 1.* The overview of the parallelized learning framework of the Deep Multivariate Probit Model.

Though equation (8) converges exponentially fast, the value of $\Pr(y_i|x_i)$ could be the magnitude of $2^{-l}$. That is, we may need to sample $O(2^{2l})$ many times to have a reasonable multiplicative error bound. To address this issue, another assertion can be proven for this sampling process using Chebyshev's inequality:

**Theorem 1** *Let $\mu \in \mathbb{R}^l$ and $\Sigma \in \mathbb{R}^{l \times l}$ be the rescaled mean and rescaled residual covariance matrix of the random variable $w^{(k)}$ in equation (7), then we have*

$$\Pr\left[\left|\frac{1}{M}\sum_{k=1}^{M}\prod_{j=1}^{l}\Phi(w_{i,j}^k) - \Pr(y_i|x_i)\right| \geq \epsilon \Pr(y_i|x_i)\right]$$

$$\leq \frac{\Phi\left(0; \begin{bmatrix} -\mu \\ -\mu \end{bmatrix}, \begin{bmatrix} \Sigma+I & \Sigma \\ \Sigma & \Sigma+I \end{bmatrix}\right)/\Phi^2(0; -\mu, \Sigma+I) - 1}{M\epsilon^2} \tag{9}$$

$$\leq \frac{\left(\frac{\Phi(0; -\mu, 2\Sigma+I)}{\Phi(0; -\mu, \Sigma+I)}\right)^2 |2\Sigma+I|^{1/2} - 1}{M\epsilon^2} \tag{10}$$

$$\leq \frac{\prod_{i=1}^{l} g(\mu_i)^2 |2\Sigma+I|^{1/2} - 1}{M\epsilon^2} \tag{11}$$

*where $g(\mu_i) = \max_x \frac{\Phi(\sqrt{2}x+\mu_i)}{\Phi(x+\mu_i)}$. See the Appendix for a more detailed proof.*

The function $g(\mu_i)$ in the theorem (1) does not have a closed form but it is a monotonous decreasing function, which converges to 1 as $\mu_i$ increases. Figure 2 is the visualization of function $g(\mu_i)$. As can be seen, the function $g(\mu_i)$ is very close to 1 when $\mu_i$ is positive. Though $g(\mu_i)$ increases exponentially with an upper bound $\sqrt{2}e^{\frac{3-2\sqrt{2}}{2}\mu_i^2}$ when $\mu_i$ is a very small negative number, the training method - maximum likelihood estimation - ensures that most $\mu_i$ are positive. In theorem (1), the equation (9) is the upper bound derived by exact analysis of the second moment of the random variable $\prod_{j=1}^{l} \Phi(w_{i,j}^k)$. Knowing there is no general closed form for the CDF of multivariate normal distribution, we further derive the equation (11) to provide an analytical upper bound.

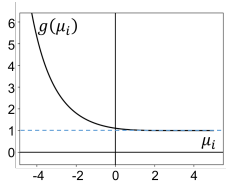

*Figure 2.* The visualization of function $g(\mu_i)$.

Though, in the worst case, the upper bounds could be exponentially large with respect to the dimensionality, this still sheds light on the convergence behavior of our sampling process. For example, if the distribution of entities is independent, then the rescaled residual covariance $\Sigma$ is a zero matrix. In that case, the variance of our sampling process is zero, so that we only need to sample once to get the exact likelihood. In more general cases, if the rescaled residual covariance $\Sigma$ is a low-rank matrix, most eigenvalues of the matrix $2\Sigma + I$ are 1, which indicates a small $|2\Sigma + I|$. According to our experiments, most eigenvalues of the rescaled residual covariance matrix $\Sigma$ are very close to 0, which supports the empirical convergence behavior of our DMVP. In the experimental section, we provide more detailed analysis in terms of the performance as well as the convergence behavior of DMVP with a low-rank residual covariance matrix, showing that the DMVP's performance only degrades significantly when the rank of the residual covariance matrix is extremely small.

Since our learning scheme is based on stochastic gradient descent, we also use the derivatives of equation (7) as the estimation of the true derivatives. The variance analysis of the derivatives of $\Pr(y_i|x_i)$, which has a similar convergence bound as theorem (1), could also be derived using the similar method. Because of space limitations, see the appendix for a more detailed proof.

## 4. Other Related Work

Multi-entity modelling problems are studied extensively under the names of multi-label classification, multi-entity embedding and structured prediction. The simplest approach is to model the distribution of each entity independently, given the contextual data, known as the *binary relevance model*. This approach is quite popular in multi-label image classification because of its simplicity and flexibility. (We chose it as a baseline for this problem.) However, this could perform poorly, particularly when certain labels are rare or some are highly correlated. Therefore, max-margin (Sarawagi & Gupta, 2008), ranking losses (Elisseeff & Weston, 2002) and embedding methods (Rudolph et al., 2016) have been used to address the correlations. Along this line of research, recent approaches (Belanger et al., 2017; Belanger & McCallum, 2016) use SSVM minimizer to optimize energy-based structured models. Those approaches mainly focus on the classification problem, in which the correlation among entities is implicit and therefore it is hard to derive the structured probabilistic distribution of entities. Our applications of DMVP, on the other hand, focus more on probabilistic modeling rather than classification. Another classic approach related to MVP is the Conditional Random Field (CRF) (Lafferty et al., 2001), which offers a general framework for structured prediction based on undirected graphical models. Instead of using correlated latent variables, CRF models the correlation among entities directly, where the joint probability of multiple outcomes is proportional to an energy function. However, optimizing CRF models suffers from the computational intractability of the partition function. To remedy this issue, (Xu et al., 2011) applied ensemble methods and (Deng et al., 2014) proposed a special CRF for problems involving specific hierarchical relations. Nevertheless, optimizing CRF models still inevitably depends on gibbs sampling for approximate inference, and has the same problem as the MCMC-based MVP models. A newly proposed ecological model, the Deep Multi-Species Embedding model (DMSE) (Chen et al., 2017), introduces deep neural networks into the classic MVP. Nevertheless, the learning methods of DMSE are also based on sequential inference such as the MCMC simulations, so that they are not easily boosted using GPUs.

The mixed-logit model (McFadden & Train, 2000) is another statistical model for analyzing discrete outcomes, whose marginal likelihood is similar to the formula of the transformation step in DMVP. However, the mixed-logit model is a general way to inject random variables into the logistic regression while the transformation in DMVP uses the auxiliary residual covariance to estimate the likelihood. Multi-Entity Dependency Learning via Conditional Variational Learning (MEDL_CVAE) uses a conditional variational auto-encoder to handle correlation between multiple entities, and is also compatible with parallelized deep structures. Despite its limitations, as discussed in the introduction, MEDL_CVAE is a state-of-the-art multi-entity modelling method and is also closely related to our DMVP model. Therefore, we chose MEDL_CVAE as the representative approach among those competitive multi-entity modelling methods and compare its performance to DMVP, in the experimental section.

# 5. Experiments

## 5.1. Datasets and Implementation Details

We evaluate our DMVP[1]. on three datasets of multi-entity modelling problems.

**eBird** is a crowd-sourced bird observation dataset collected from the successful citizen science project *eBird* (Munson et al., 2012). One record in this dataset is referred to as a checklist in which the bird observer records all the species he/she detects as well as the time and the geographical location of the observational site. Crossed with the National Land Cover Dataset for the U.S. (NLCD) (Homer et al., 2015), we obtain a 15-dimensional feature vector for each observational site which describes the landscape composition with respect to 15 different land types such as water, forest, etc. We also collect the satellite images for each observation site by matching the geographical location of the observational site to Google Earth[2]. Each satellite image covers an area of 12.3km$^2$ near the observation site and has 256×256 pixels. The dataset for this experiment is formed by picking all the observation checklists from the Bird Conservation Region (BCR) 13 (Committee et al., 2000) in the last two weeks of May from 2004 to 2014, which contains 50,949 observations. We choose the top 100 most frequently observed birds as the target species which cover over 95% of the records in our dataset.

**Amazon** is the Amazon rainforest landscape satellite image dataset collected for Amazon rainforest landscape analysis,[3] in which raw images were derived from Planet's full-frame analytic scene products using 4-band satellites in sun-synchronous orbit and International Space Station orbit. The organizers used Planet's visual product processor to transform raw images into 3-band 256x256-pixel jpg format. The *Amazon* contains a total of 34,431 samples and each sample in this dataset contains a satellite image chip covering an area of 0.9 km$^2$ in Amazon rainforest. The chips were analyzed using the Crowd Flower[4] platform to obtain a ground-truth composition of the landscape. There are 17 different labels for each satellite image chip, which represent a reasonable subset of phenomena of interest in the Amazon basin such as atmospheric conditions, common land cover phenomena, and land use phenomena.

**NUS-WIDE-LITE** is a light version of the *NUS-WIDE* datasets collected by the National University of Singapore

---

---

| Dataset | eBird | Amazon | NUS |
|---|---|---|---|
| #Training Set | 40759 | 27545 | 44493 |
| #Validation Set | 5095 | 3443 | 5561 |
| #Test Set | 5095 | 3443 | 5561 |
| #Entities | 100 | 17 | 81 |

*Table 1.* the statistics of the *eBird* and the *Amazon* dataset

(Chua et al., July 8-10, 2009), which contains 55,615 samples and each sample is the low-level features (such as wavelet texture, histogram, correlogram, etc) of the real-world web image associated with tags from Flicker. The 81 tags represent 81 different concepts related to the web images, such as the concepts related to the objects in the image (dog, cat, building, etc) and the concepts of the background (clouds, sunset, etc). For the ease of presentation, we use **NUS** to denote this dataset.

We randomly split the datasets into three parts for training, validation, and testing. The details of the three datasets are listed in table 1.

## 5.2. Performance Analysis of the DMVP on Multi-Entity Modelling Problems

We compare the proposed DMVP with baseline models from three different groups. The first group, which we refer to as *conditional independent model* (CIM), assumes independence among entities, conditioned on the contextual data. Within this group, we chose different models based on the type of the input features. For example, when the input features are images, we choose to use convolutional neural networks (CNN), while we use the multi-layer fully connected neural network (MLP) for one-dimensional feature inputs. For the sake of fairness, the structure of CIM as well as the feature networks in other baseline models are always the same as the feature network of DMVP. More specifically, for the data resources of low-level features, such as the NLCD features of **eBird** dataset and the **NUS** dataset, we use a 4-layer fully connected neural network with hidden units of size 128, 256, 256, $l$, where the activation function of the first 3 layers is ReLU (Nair & Hinton, 2010) and there is no activation function in the last layer. For the image data resources, we use a CNN similar to the Alexnet (Krizhevsky et al., 2012) with some minor adjustments. The second group is the *previously proposed Multivariate Probit Model*, which can also model correlations among entities, but uses different inference methods. Within this group, we chose the Deep Multi-Species Embedding (DMSE) model (Chen et al., 2017), a gradient-based MVP model, which uses the numerical computing method proposed by (Genz, 1992) to estimate the likelihood and a MCMC-based method to estimate the gradients. This model represents a wide class of MCMC-based multivariate probit models while further improving the classic MVP by taking advantages of the flexibility of deep neural networks to obtain useful feature extractions. Nevertheless, its training process involves MCMC approaches as well as the sequential importance sampling, and therefore cannot be integrated on GPUs. For the last group, we chose the `MEDL_CVAE`(Tang et al., 2017) model, which is a *state-of-the-art multi-entity modelling approach* proposed recently. This model uses conditional variational auto-encoder to handle correlation between multiple entities, in which it approximates the joint likelihood by its variational lower bound.

Because we study multi-entity modelling problems, in our experiments, we use Negative Joint Distribution Log-likelihood (Neg.JLL) as the indicator of a model's performance: $-\frac{1}{N}\sum_{i=1}^{N}\log\Pr(y_i|x_i)$, where $N$ is the number of samples in the test set. Based on the theorem (1) we obtain 1,000,000 samples from the residual multivariate normal distribution for testing DMVP's performance, which is sufficient to guarantee the accuracy of the estimation. However, for the training, DMVP empirically converges well with only 100 samples.

All the training and testing process of our DMVP and other baseline models, which are compatible with the GPUs, are performed on one NVIDIA Quadro P4000 GPU with 8GB memory. The training and testing process for the DMSE model is performed on Intel(R) Core(TM) i7-7700K CPU@4.20Gz with 8 cores. Since the bottleneck of the DMSE model is on the MCMC sampling, which could not be parallelized trivially, additional cores do not improve the wall-clock time significantly. The whole training process lasts 200 epochs, using the batch size of 128, Adam optimizer (Kingma & Ba, 2014) with learning rate of $10^{-4}$ and utilizing batch normalization (Ioffe & Szegedy, 2015), 0.5 dropout rate (Srivastava et al., 2014) and early stopping to accelerate the training process and to prevent overfitting for not only DMVP but all baseline models.

Table 2 shows the average performance of DMVP as well as other baseline models on the 3 datasets (4 different type of input features) in terms of the negative joint log-likelihood (Neg.JLL) and the wall-clock time of training.[5] There are multiple key results in Table 2: **(1)** By comparing the Neg.JLL of the conditional independent model (CIM) with other models, one can observe **significant advantages of modelling the correlations among entities.** **(2) DMVP trains more than 100 times faster than the MCMC-based DMSE model** in terms of the wall-clock time. This huge gap between DMVP and DMSE is due not only to the parallelization but also to the advantage of sampling from an explicit distribution. For DMVP, empirically we only need to sample 100 samples per data point to converge very well and every sample here is an unbiased esti-mation of the joint likelihood. However, in DMSE, we need to burn every 1000 intermediate samples to merely get one quasi-unbiased sample from the implicit distribution, which is not cost-efficient. What's more, the high-resolution image data resources are way beyond the capacity of the MCMC-based method, where the DMSE model cannot reach a reasonable performance after 2-day-long training. **(3)** In terms of the Neg.JLL, **DMVP outperforms all baseline models** including the competitive `MEDL_CVAE` model, which is compatible with deep neural networks and also models the correlations among entities. There are two reasons of why DMVP outperforms `MEDL_CVAE`: (i) DMVP directly learns the joint likelihood while `MEDL_CVAE` approximates the joint likelihood by optimizing its variational lower bound. (2) there is a KL-vanishment issue, which is notorious in all applications based on variational autoencoder, in the training of variational lower bound that hampers the performance of the `MEDL_CVAE` model.

---

[5]We thank the authors of (Tang et al., 2017) and (Chen et al., 2017) for sharing the codes.

| eBird-NLCD | | | | |
|---|---|---|---|---|
| **Method** | **CIM** | **DMSE** | `MEDL_CVAE` | **DMVP** |
| wall-clock time (mins) | **2** | 1200 | 10 | 10 |
| Neg.JLL | 34.96 | 30.53 | 30.86 | **29.68** |
| eBird-Images | | | | |
| **Method** | **CIM** | **DMSE** | `MEDL_CVAE` | **DMVP** |
| wall-clock time (mins) | **820** | >3000 | 847 | 843 |
| Neg.JLL | 34.14 | N/A | 33.68 | **28.26** |
| Amazon | | | | |
| **Method** | **CIM** | **DMSE** | `MEDL_CVAE` | **DMVP** |
| wall-clock time (mins) | **484** | >3000 | 502 | 495 |
| Neg.JLL | 1.70 | N/A | 1.64 | **1.50** |
| NUS-WIDE-LITE | | | | |
| **Method** | **CIM** | **DMSE** | `MEDL_CVAE` | **DMVP** |
| wall-clock time (mins) | **4** | 1410 | 12 | 12 |
| Neg.JLL | 6.17 | 5.76 | 5.82 | **5.73** |

*Table 2.* Comparison of various methods on 3 datasets ( 4 different input features) in terms of the Negative Joint Log-likelihood (the smaller the better ) and the wall-clock time.

### 5.3. Empirical Analysis of the DMVP's Convergence Behavior

In Section 3.2 we provide the theoretical upper bound of the DMVP's convergence behavior. Based on the theorem (1), one way to reduce the sampling variance is to assume the low-rank property of the residual covariance matrix. Therefore, we conducted the empirical analysis of the DMVP's performance as well as the convergence behavior on three datasets with residual covariance matrix of different rank. Based on the theorem (1), we use the numerators of both equation (9) and equation (11) to indicate the convergence rate, where the former is a tighter bound without an analytic form and the latter is the theoretical upper bound. Though

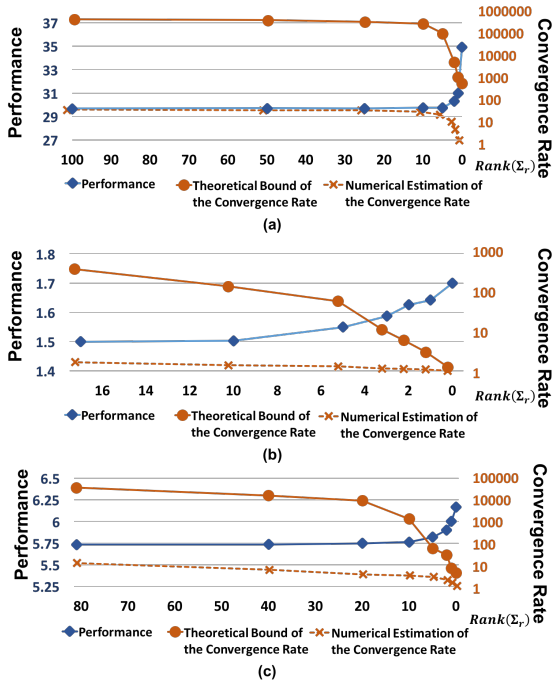

*Figure 3.* The analysis of DMVP's performance and the convergence behavior on three datasets with respect to low-rank residual covariance matrix. The performance is indicated by Neg.JLL and the convergence rate are measured using both the theoretical bound derived from equation (11) as well as the numerical estimation of the tighter bound derived from equation (9). (For both of them, the smaller the better.) As the rank of $\Sigma_r$ goes lower, the DMVP converges better while the performance of DMVP only degrades significantly when the rank of $\Sigma_r$ is extremely low. The subplots (a) (b) (c) correspond to **eBird**, **Amazon** and **NUS** respectively.

the tighter bound (equation (9)) does not have an analytic form, we could show the value estimated using the numerical method proposed in (Genz, 1992). What's more, because the value of the indicators derived from equation (11) and equation (9) vary across data points over time, we pick the median at the end of the training as the representative. We implement the constraint of $rank(\Sigma_r)$ by restricting the dimensionality of $\Sigma_r^{1/2}$, i.e., $\Sigma_r^{1/2} \in \mathbb{R}^{l \times k}$, where $k \leq l$. Figure (3) show the experimental results conducted on three datasets. Because of the similarity, for the **eBird** dataset, we only show the analysis using NLCD features. One observation from Figure (3) is that the theoretical bound is way looser than the numerical estimation of the tighter bound, which is actually closer to the empirical results. In our experiments, DMVP converges well in all datasets using only 100 samples. Nevertheless, the theoretical bound still sheds light on the convergence behavior of DMVP. One can see, as we restrict the rank of the residual covariance matrix to be lower, both the theoretical bound and the numerical estimation of the convergence rate become better while the performance of DMVP only degrades significantly

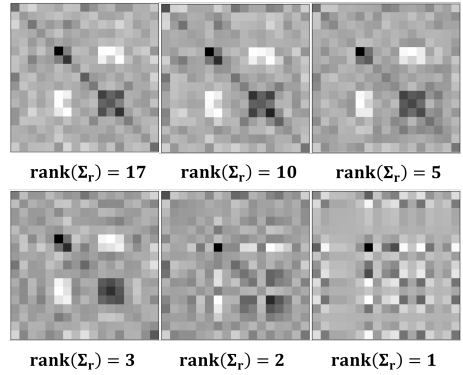

| rank($\Sigma_r$) = 17 | rank($\Sigma_r$) = 10 | rank($\Sigma_r$) = 5 |

| rank($\Sigma_r$) = 3 | rank($\Sigma_r$) = 2 | rank($\Sigma_r$) = 1 |

*Figure 4.* The visualization of the residual covariance matrix ($\Sigma_r$) on the **Amazon** dataset, with $\Sigma_r$ of different ranks, which capture the correlations in different resolutions. The pattern of $\Sigma_r$ only degenerates with extremely low ranks. (less or equal to 3)

when the rank of $\Sigma_r$ is extremely small. The reason behind this phenomenon is that the rank of $\Sigma_r$ actually describes the the resolution of how fine-grained DMVP models the residual covariance. Therefore, it is possible to approximate a full-rank matrix by a low-rank matrix with minimal discrepancy. As an example, the Figure 4 is the heatmap of the residual covariance matrix on **Amazon** dataset with rank from full-rank to rank-1. (Because of the space limitation, we only show the covariance heatmap of **Amazon** datasets.) One can see, the pattern of residual covariance does not change too much until the resolution is extremely low. These facts are consistent with the empirical results of learning DMVP with full-rank residual covariance matrix, where most eigenvalues of $\Sigma_r$ are very close to zero. Based on these observations, we can naturally balance the computational complexity and the predictive performance of DMVP by tuning the resolution. This provides the potential benefits of using DMVP to analyze large scale multi-entity correlation with low-rank constraints.

## 6. Conclusion

In this paper, we propose an end-to-end learning scheme for DMVP, in which we propose an efficient parallel sampling process to integrate DMVP with various GPU-boosted deep neural networks. Tested on three real-world applications of multi-entity modelling, we show that DMVP trains 100x faster than previous MCMC-based methods, captures rich correlations among entities, and consistently performs better than previous models. We further provide both theoretical and empirical analysis of DMVP's convergence behavior, revealing the benefits of balancing the computational complexity and the predictive performance by restricting the rank of the residual covariance matrix. Future directions include exploring the potential of applying DMVP on large scale correlation analysis with the low-rank residual covariance constraint.

## Acknowledgement

We are thankful to thousands of eBird participants, and the Cornell Lab of Ornithology. This research was supported by National Science Foundation (Grant Number 0832782, 1522054, 1059284, 1356308), and ARO grant W911-NF-14-1-0498.

## References

Ashford, J. and Sowden, R. Multi-variate probit analysis. *Biometrics*, pp. 535–546, 1970.

Belanger, D. and McCallum, A. Structured prediction energy networks. In *International Conference on Machine Learning*, pp. 983–992, 2016.

Belanger, D., Yang, B., and McCallum, A. End-to-end learning for structured prediction energy networks. *arXiv preprint arXiv:1703.05667*, 2017.

Blackwell, D. Conditional expectation and unbiased sequential estimation. *The Annals of Mathematical Statistics*, pp. 105–110, 1947.

Chen, D., Xue, Y., Chen, S., Fink, D., and Gomes, C. Deep multi-species embedding. In *IJCAI*, 2017.

Chib, S. and Greenberg, E. Analysis of multivariate probit models. *Biometrika*, 85(2):347–361, 1998.

Chua, T.-S., Tang, J., Hong, R., Li, H., Luo, Z., and Zheng, Y.-T. Nus-wide: A real-world web image database from national university of singapore. In *Proc. of ACM Conf. on Image and Video Retrieval (CIVR'09)*, Santorini, Greece., July 8-10, 2009.

Cody, W. J. Rational chebyshev approximations for the error function. *Mathematics of Computation*, 23(107): 631–637, 1969.

Committee, U. N. et al. North american bird conservation initiative: Bird conservation region descriptions, a supplement to the north american bird conservation initiative bird conservation regions map. 2000.

Deng, J., Ding, N., Jia, Y., Frome, A., Murphy, K., Bengio, S., Li, Y., Neven, H., and Adam, H. Large-scale object classification using label relation graphs. In *ECCV*, 2014.

Elisseeff, A. and Weston, J. A kernel method for multi-labelled classification. In *Advances in neural information processing systems*, pp. 681–687, 2002.

Genz, A. Numerical computation of multivariate normal probabilities. *Journal of computational and graphical statistics*, 1992.

Gomes, C. P. Computational sustainability: Computational methods for a sustainable environment, economy, and society. *The Bridge*, 39(4):5–13, 2009.

Homer, C., Dewitz, J., Yang, L., Jin, S., Danielson, P., Xian, G., Coulston, J., Herold, N., Wickham, J., and Megown, K. Completion of the 2011 national land cover database for the conterminous united states–representing a decade of land cover change information. *Photogrammetric Engineering & Remote Sensing*, 2015.

Ioffe, S. and Szegedy, C. Batch normalization: Accelerating deep network training by reducing internal covariate shift. In *International Conference on Machine Learning*, pp. 448–456, 2015.

Jeliazkov, I. and Hee Lee, E. Mcmc perspectives on simulated likelihood estimation. In *Maximum simulated likelihood methods and applications*, pp. 3–39. Emerald Group Publishing Limited, 2010.

Kingma, D. and Ba, J. Adam: A method for stochastic optimization. *arXiv preprint arXiv:1412.6980*, 2014.

Krizhevsky, A., Sutskever, I., and Hinton, G. E. Imagenet classification with deep convolutional neural networks. In *Advances in neural information processing systems*, pp. 1097–1105, 2012.

Lafferty, J. D., McCallum, A., and Pereira, F. C. N. Conditional random fields: Probabilistic models for segmenting and labeling sequence data. In *Proceedings of the Eighteenth International Conference on Machine Learning*, ICML '01, 2001.

MacKenzie, D. I., Bailey, L. L., Nichols, J., et al. Investigating species co-occurrence patterns when species are detected imperfectly. *Journal of Animal Ecology*, pp. 546–555, 2004.

Magid, A. R. *Lectures on differential Galois theory*. Number 7. American Mathematical Soc., 1994.

Mandt, S., Wenzel, F., Nakajima, S., Cunningham, J., Lippert, C., and Kloft, M. Sparse probit linear mixed model. *Machine Learning*, 106(9-10):1621–1642, 2017.

McFadden, D. and Train, K. Mixed mnl models for discrete response. *Journal of applied Econometrics*, pp. 447–470, 2000.

Munson, M. A., Webb, K., Sheldon, D., Fink, D., Hochachka, W. M., Iliff, M., Riedewald, M., Sorokina, D., Sullivan, B., Wood, C., et al. The ebird reference dataset, version 4.0. 2012.

Nair, V. and Hinton, G. E. Rectified linear units improve restricted boltzmann machines. In *Proceedings of the 27th*

*international conference on machine learning (ICML-10)*, pp. 807–814, 2010.

Nam, J., Kim, J., Mencía, E. L., Gurevych, I., and Fürnkranz, J. Large-scale multi-label text classificationrevisiting neural networks. In *Joint european conference on machine learning and knowledge discovery in databases*, pp. 437–452. Springer, 2014.

Rudolph, M., Ruiz, F., Mandt, S., and Blei, D. Exponential family embeddings. In *Advances in Neural Information Processing Systems*, 2016.

Sarawagi, S. and Gupta, R. Accurate max-margin training for structured output spaces. In *Proceedings of the 25th international conference on Machine learning*, pp. 888–895. ACM, 2008.

Song, X.-Y. and Lee, S.-Y. A multivariate probit latent variable model for analyzing dichotomous responses. *Statistica Sinica*, pp. 645–664, 2005.

Srivastava, N., Hinton, G. E., Krizhevsky, A., Sutskever, I., and Salakhutdinov, R. Dropout: a simple way to prevent neural networks from overfitting. *Journal of machine learning research*, 15(1):1929–1958, 2014.

Tabet, A. *Bayesian inference in the multivariate probit model*. PhD thesis, University of British Columbia, 2007.

Tang, L., Xue, Y., Chen, D., and Gomes, C. Multi-entity dependence learning with rich context via conditional variational auto-encoder. In *AAAI*, 2017.

Wang, J., Yang, Y., Mao, J., Huang, Z., Huang, C., and Xu, W. Cnn-rnn: A unified framework for multi-label image classification. In *Proceedings of the IEEE Conference on Computer Vision and Pattern Recognition*, pp. 2285–2294, 2016.

Xu, X.-S., Jiang, Y., Peng, L., Xue, X., and Zhou, Z.-H. Ensemble approach based on conditional random field for multi-label image and video annotation. In *Proceedings of the 19th ACM International Conference on Multimedia*, 2011.

Young, G., Valdez, E. A., and Kohn, R. Multivariate probit models for conditional claim-types. *Insurance: Mathematics and Economics*, 44(2):214–228, 2009.

# 7. Appendix

**Theorem 1** *Let $\mu \in R^l$ and $\Sigma \in R^{l \times l}$ be the rescaled mean and the rescaled residual covariance matrix of the random variable $w^{(k)}$ in the equation (7) of the main text, then we have*

$$
\Pr\left[ \left| \frac{1}{M} \sum_{k=1}^{M} \prod_{j=1}^{l} \Phi(w_{i,j}^{(k)}) - \Pr(y_i|x_i) \right| \geq \epsilon \Pr(y_i|x_i) \right]
$$

$$
\leq \frac{\Phi\left(0; \begin{bmatrix} -\mu \\ -\mu \end{bmatrix} \cdot \begin{bmatrix} \Sigma + I & \Sigma \\ \Sigma & \Sigma + I \end{bmatrix}\right) - \Phi^2(0; -\mu, \Sigma + I)}{M \Phi^2(0; -\mu, \Sigma + I) \epsilon^2} \tag{12}
$$

$$
\leq \frac{\left( \frac{\Phi(0; -\mu, 2\Sigma+I)}{\Phi(0; -\mu, \Sigma+I)} \right)^2 |2\Sigma + I|^{1/2} - 1}{M \epsilon^2} \tag{13}
$$

$$
\leq \frac{\prod_{i=1}^{l} g(\mu_i)^2 |2\Sigma + I|^{1/2} - 1}{M \epsilon^2} \tag{14}
$$

*where $g(\mu_i) = \max_x \frac{\Phi(\sqrt{2}x+\mu_i)}{\Phi(x+\mu_i)}$. The function $g(\mu_i)$ does not have a closed form but it is a monotonous decreasing function, which converges to 1 as $\mu_i$ increases.*

*Proof.* For the ease of expression, we omit the subscripts related to $i$-th data point in our proof. Without loss of generality, we can also assume the diagonal matrix $V$ is an indentity matrix. Defining $\Pr(y|w) = \prod_{j=1}^{n} \Phi(w_j)$, $\Pr(y|x) = E_{w \sim N(\mu, \Sigma)}[\Pr(y|w)]$. We prove this convergence bound by analysing the first and second moment of random variable $\Pr(y|w)$.

$$
E_w[\Pr(y|w)] = \int_w \prod_{j=1}^{n} \Phi(w_j) Pr_w(w) \mathrm{d}w
$$

$$
= \int_w Pr_z(z \preceq w|w) Pr_w(w) \mathrm{d}w
$$

$$
= Pr_{z,w}(z \preceq w)
$$

$$
= Pr_{z,w}(z - w \preceq 0) \tag{15}
$$

Here $z \sim N(0, I)$ and $a \preceq b$ means $\forall a_i \leq b_i$

Since $z$ is subject to multivariate gaussian distribution, $z - w$ is still a multivariate gaussian random variable, which is subject to $N(-\mu, \Sigma + I)$. Thus, $\Pr(y|x) = E_w[\Pr(y|w)] = \Phi(0; -\mu, \Sigma + I)$. ($\Phi(\cdot)$ denotes the cumulative function of multivariate gaussian distribution.)

Similarly, we can derive that

$$
E[\Pr(y|w)^2] = \Pr(z_1 \preceq w \wedge z_2 \preceq w)
$$

$$
= \Pr\left( \begin{bmatrix} z_1 \\ z_2 \end{bmatrix} \preceq \begin{bmatrix} r \\ r \end{bmatrix} \right)
$$

$$
= \Phi\left( 0; \begin{bmatrix} -\mu \\ -\mu \end{bmatrix}, \begin{bmatrix} \Sigma + I & \Sigma \\ \Sigma & \Sigma + I \end{bmatrix} \right)
$$

Let $B = \begin{bmatrix} \Sigma + I & \Sigma \\ \Sigma & \Sigma + I \end{bmatrix}$, we have $|B| = \left| det\left( \begin{bmatrix} 2\Sigma + I & \Sigma \\ 0 & I \end{bmatrix} \right) \right| = |2\Sigma + I|$. Since $\Sigma$ is a positive definite matrix, we can decompose $\Sigma = UDU^T$, where $U$ is

an orthogonal matrix and $D$ is a diagonal matrix. Similarly, we can decompose

$$B^{-1} = \begin{bmatrix} U & 0 \\ 0 & U \end{bmatrix} \begin{bmatrix} (2D+I)^{-1}(D+I) & -(2D+I)^{-1}D \\ -(2D+I)^{-1}D & (2D+I)^{-1}(D+I) \end{bmatrix} \begin{bmatrix} U^T & 0 \\ 0 & U^T \end{bmatrix}$$

Let $x_1, x_2 \in R^l$, $y_1 = U^T(x_1 + \mu)$, $y_2 = U^T(x_1 + \mu)$ and $D = diag(d_1, ..., d_l)$, then we have,

$$E[\Pr(y|r)^2] = \Phi\left(0; \begin{bmatrix} -\mu \\ -\mu \end{bmatrix}, \begin{bmatrix} \Sigma + I & \Sigma \\ \Sigma & \Sigma + I \end{bmatrix}\right)$$

$$= \frac{1}{(2\pi)^l |B|^{1/2}} \int_{(-\infty, 0]^l} e^{-\frac{1}{2}(\sum_{i=1}^{l}(y_{1,i}^2 + y_{2,i}^2)\frac{d_i+1}{2d_i+1} - 2\sum_{i=1}^{l} y_{1,i} y_{2,i} \frac{d_i}{2d_i+1})} dx_1 dx_2$$

$$\leq \frac{1}{(2\pi)^l |B|^{1/2}} \int_{(-\infty, 0]^l} e^{-\frac{1}{2}(\sum_{i=1}^{l}(y_{1,i}^2 + y_{2,i}^2)\frac{1}{2d_i+1})} dx_1 dx_2$$

$$= |2\Sigma + I|^{1/2} \Phi\left(0; \begin{bmatrix} -\mu \\ -\mu \end{bmatrix}, \begin{bmatrix} 2\Sigma + I & 0 \\ 0 & 2\Sigma + I \end{bmatrix}\right)$$

Thus,

$$E[\Pr(y|r)^2]^{1/2} \leq |2\Sigma + I|^{1/4} \Phi(0; -\mu, 2\Sigma + I)$$

Using the inverse transformation in equation (15), we have

$$\Phi(0; -\mu, 2\Sigma + I)$$
$$= \frac{1}{(2\pi)^{1/2}|2\Sigma|^{1/2}} \int \prod \Phi(x) e^{\frac{1}{4}(x-\mu)^T \Sigma^{-1}(x-\mu)} dx$$
$$= \frac{1}{(2\pi)^{1/2}|\Sigma|^{1/2}} \int \prod \Phi(\sqrt{2}y + \mu_i) e^{\frac{1}{2}y^T \Sigma^{-1} y} dy$$

$$(16)$$

Let $g(\mu_i) = \max_x \frac{\Phi(\sqrt{2}x + \mu_i)}{\Phi(x + \mu_i)}$, then we have

$$\Phi(0; -\mu, 2\Sigma + I)$$
$$= \frac{1}{(2\pi)^{1/2}|\Sigma|^{1/2}} \int \prod \Phi(\sqrt{2}y + \mu) e^{\frac{1}{2}y^T \Sigma^{-1} y} dy$$
$$\leq \frac{\prod_{i=1}^{l} g(\mu_i)}{(2\pi)^{1/2}|\Sigma|^{1/2}} \int \prod \Phi(y + \mu) e^{\frac{1}{2}y^T \Sigma^{-1} y} dy$$
$$= \prod_{i=1}^{l} g(\mu_i) \Phi(\mu|\Sigma + I)$$
$$= \prod_{i=1}^{l} g(\mu_i) \Pr(y|x)$$

Therefore,

$$E[\Pr(y|w)^2]^{1/2} \leq |2\Sigma + I|^{1/4} \Phi(0; -\mu, 2\Sigma + I)$$
$$\leq |2\Sigma + I|^{1/4} \prod_{i=1}^{l} g(\mu_i) \Phi(0; -\mu, \Sigma + I)$$

Using the Chebyshev's inequality, we have

$$\Pr[|\frac{1}{M} \sum_{k=1}^{M} \prod_{j=1}^{l} \Phi(w_{i,j}^{(k)}) - \Pr(y_i|x_i)| \geq \epsilon \Pr(y_i|x_i)]$$

$$= \Pr[|\frac{1}{M} \sum_{k=1}^{M} \Pr(y_i|w_i^{(k)}) - \Pr(y_i|x_i)| \geq \epsilon \Pr(y_i|x_i)]$$

$$= \Pr[|\frac{1}{M} \sum_{k=1}^{M} \Pr(y_i|w_i^{(k)}) - \Pr(y_i|x_i)|^2 \geq \epsilon^2 \Pr(y_i|x_i)^2]$$

$$\leq \frac{E[(\frac{1}{M} \sum_{k=1}^{M} \Pr(y_i|w_i^{(k)}) - \Pr(y_i|x_i))^2]}{\epsilon^2 \Pr(y_i|x_i)^2}$$

$$= \frac{\prod_{i=1}^{l} g^2(\mu_i) |2\Sigma + I|^{1/2} - 1}{M\epsilon^2} \quad \blacksquare$$

The function $g(\mu_i)$ does not have a closed form but it is a monotonous decreasing function, which converges to 1 as $\mu_i$ increases. The figure (5) is the visualization of function $g(\mu_i)$. As you see, the function $g(\mu_i)$ is very close to 1 when $\mu_i$ is positive. The following lemma provides a more analytical upper bound for function $g(\mu_i)$.

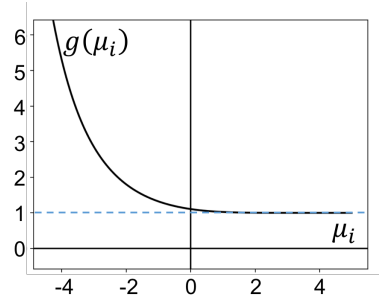

Figure 5. The visualization of function $g(\mu_i)$.

**Lemma 1** *For any $y$, $\Phi(\sqrt{2}y + \mu) \leq g(\mu)\Phi(y + \mu)$, where*

$$g(\mu) \leq \begin{cases} \sqrt{2} e^{-\frac{3 - 2\sqrt{2}}{2}\mu^2} & \text{if } \mu < 0 \\ 1.182 & \text{if } \mu \geq 0 \end{cases}$$

*Proof.* $\frac{\Phi(\sqrt{2}y + \mu)}{\Phi(y + \mu)}$ achieves the maximum when its derivative is equal to zero, i.e.,

$$\left(\frac{\Phi(\sqrt{2}y + \mu)}{\Phi(y + \mu)}\right)' = 0 \Longrightarrow$$

$$\frac{\frac{1}{\sqrt{2\pi}}(\sqrt{2}e^{-\frac{1}{2}(\sqrt{2}y + \mu)^2}\Phi(y + \mu) - e^{-\frac{1}{2}(y + \mu)^2}\Phi(\sqrt{2}y + \mu)}{\Phi^2(y + \mu)} = 0$$

$$\Longrightarrow \frac{\Phi(\sqrt{2}y + \mu)}{\Phi(y + \mu)} = \sqrt{2}e^{-\frac{1}{2}(y^2 + 2(\sqrt{2}-1)\mu y)}$$

Since $\Phi(x)$ is a monotonic increasing function, $max_y \sqrt{2} e^{-\frac{1}{2}(y^2 + 2(\sqrt{2}-1)\mu y)} = \sqrt{2} e^{\frac{3-2\sqrt{2}}{2}\mu^2}$ when $\mu < 0$. Similarly, when $\mu \geq 0$, we know $y^* = argmax_y \frac{\Phi(\sqrt{2}y+\mu)}{\Phi(y+\mu)} \geq 0$. Thus, $\Phi(y^* + \mu) \geq \frac{1}{2}$. By analysing the maximal value of $\Phi(\sqrt{2}y + \mu) - \Phi(y + \mu)$ as well as the fact that $\Phi(\sqrt{2}y + \mu) - \Phi(y + \mu) \leq (\sqrt{2}-1)y * \frac{1}{\sqrt{2\pi}} e^{-\frac{1}{2}(y+\mu)^2}$, we could know that $\Phi(\sqrt{2}y + \mu) - \Phi(y + \mu) \leq 0.091$. That is,

$$g(\mu) \leq \begin{cases} \sqrt{2} e^{\frac{3-2\sqrt{2}}{2}\mu^2} & \text{if } \mu < 0 \\ 1.182 & \text{if } \mu \geq 0 \end{cases}$$

.

**Theorem 2** *Let $\mu \in R^l$ and $\Sigma \in R^{l \times l}$ be the rescaled mean and rescaled residual covariance matrix of the random variable $w^{(k)}$ in equation (7) of the main text, we have*

$$\Pr\left[\left|\frac{\partial \frac{1}{M}\sum_{k=1}^{M}\prod_{j=1}^{l}\Phi(w_{i,j}^k)}{\partial \mu_i} - \frac{\partial \Pr(y_i|x_i)}{\partial \mu_i}\right| \geq \epsilon \frac{\partial \Pr(y_i|x_i)}{\partial \mu_i}\right]$$

$$\leq \frac{e^{\frac{\mu_i^2}{2(\Sigma_{i,i}+1)}}(\Sigma_{i,i}+1)\lambda_{max}\prod_{j\neq i}^{l}g(\mu_j')^2|2\Sigma+I|^{1/2}-1}{M\epsilon^2} \quad (17)$$

*Here $\lambda_{max}$ denotes the largest eigenvalue of $\Sigma$ and $\mu' = \mu - \frac{\mu_i}{v+1}\Sigma^{1/2}b_i$. ($b_i$ denotes the $i$-th row of $\Sigma_{1/2}$.)*

*Proof.* For the ease of symbolism, we omit all the subscript $i$ related to the index of $i$-th data point. For any $1 \leq i \leq l$,

$$\frac{\partial \Pr(y|x)}{\partial \mu_i} = E_{w \sim N(\mu, \Sigma)}\left[\frac{\partial \prod_{j=1}^{l}\Phi(w_j)}{\partial \mu_i}\right]$$

$$= \int \prod_{j\neq i}^{l}\Phi(w_j) * \phi(w_i)\phi(w|\mu, \Sigma)dw$$

$$= \int \prod_{j\neq i}^{l}\Phi(\Sigma_j^{1/2}x + \mu_j) * \phi(\Sigma_i^{1/2}x + \mu_i)\phi(x|0, I)dx$$

Let $B = \Sigma^{1/2}$ and let $b_j$ denote the $j$-th row of $B$.

$$= \int \prod_{j\neq i}^{l}\Phi(b_j^T x + \mu_j) * \phi(b_i^T x + \mu_i)\phi(x|0, I)dx$$

let $v = b_i^T b_i = \Sigma_{i,i}$ and $C = I - \frac{b_i b_i^T}{v+1}$ ($C^{-1} = I + b_i b_i^T$).

$$= \phi(\frac{\mu_i}{v+1}) * |C|^{1/2} \int \prod_{j\neq i}^{l}\Phi(b_j^T x + \mu_j) * \phi(x| -\frac{\mu_i}{v+1}b_i, C)dx$$

$$= \phi(\frac{\mu_i}{v+1}) * |C|^{1/2} * \Pr(\forall j \neq i, z_j \leq b_j^T x + \mu_j)$$

(where $x \sim N(-\frac{\mu_i}{v+1}b_i, C)$ and $z \sim N(0, I)$.)

$$= \phi(\frac{\mu_i}{v+1}) * |C|^{1/2} * \Pr(z \preceq w)$$

(where $w \sim N(\mu_{-i} - \frac{\mu_i}{v+1}B_{-i}b_i, B_{-i}CB_{-i}^T)$, $\mu_{-i} \in R^{l-1}$ denotes the vector derived from $\mu$ by eliminating the $i$-th entry. $B_{-i} \in R^{l-1 \times l}$ denotes the matrix derived from $B$ by eliminating the $i$-th row.)

Thus, using the transformation above, we can transform the derivative in terms of $\mu_i$ into the form similar to theorem (1). Because $B_{-i}CB_{-i}^T = B_{-i}B_{-i}^T - \frac{(B_{-i}b_i)(B_{-i}b_i)^T}{v+1}$, where $B_{-i}B_{-i}^T$ is a principal submatrix of $\Sigma$, whose eigenvalues are interlaced with the eigenvalues of $\Sigma$, and $\frac{(B_{-i}b_i)(B_{-i}b_i)^T}{v+1}$ is a rank-1 matrix, we have $|2B_{-i}CB_{-i}^T + I| \leq |2\Sigma + I| * \lambda_{max}$.

In terms of the second moment of the derivative of $\mu_i$, we have,

$$E_{w \sim N(\mu, \Sigma)}\left[\left(\frac{\partial \prod_{j=1}^{l}\Phi(w_j)}{\partial \mu_i}\right)^2\right]$$

$$= \int \prod_{j\neq i}^{l}\Phi^2(\Sigma_j^{1/2}x + \mu_j) * \phi^2(\Sigma_i^{1/2}x + \mu_i)\phi(x|0, I)dx$$

$$\leq \int \prod_{j\neq i}^{l}\Phi^2(\Sigma_j^{1/2}x + \mu_j) * \phi(\Sigma_i^{1/2}x + \mu_i)\phi(x|0, I)dx$$

$$= \phi(\frac{\mu_i}{v+1}) * |C|^{1/2} \int \prod_{j\neq i}^{l}\Phi^2(b_j^T x + \mu_j) * \phi(x| -\frac{\mu_i}{v+1}b_i, C)dx$$

$$= \phi(\frac{\mu_i}{v+1}) * |C|^{1/2} * \Pr(z^1 \preceq w \wedge z^2 \preceq w)$$

Here we use the same notation as the proof above.

Using the similar trick as theorem (1), we have

$$\Pr\left[\left|\frac{\partial \frac{1}{M}\sum_{k=1}^{M}\prod_{j=1}^{l}\Phi(w_{i,j}^k)}{\partial \mu_i} - \frac{\partial \Pr(y_i|x_i)}{\partial \mu_i}\right| \geq \epsilon \frac{\partial \Pr(y_i|x_i)}{\partial \mu_i}\right]$$

$$\leq \frac{e^{\frac{\mu_i^2}{2(v+1)}}|C^{-1}|\lambda_{max}\prod_{j\neq i}^{l}g(\mu_j')^2|2\Sigma+I|^{1/2}-1}{M\epsilon^2}$$

$$\leq \frac{e^{\frac{\mu_i^2}{2(\Sigma_{i,i}+1)}}(\Sigma_{i,i}+1)\lambda_{max}\prod_{j\neq i}^{l}g(\mu_j')^2|2\Sigma+I|^{1/2}-1}{M\epsilon^2}$$

Here $\mu' = \mu - \frac{\mu_i}{v+1}\Sigma^{1/2}b_i$.

In this way, we bound the convergence of the derivatives in terms of $\mu$, so that the derivatives in term of the parameters in feature network can be derived by chain rule. However, because the derivatives of $\Sigma^{1/2}$ could be negative or zero, we can not apply the Chebyshev's inequality to have a similar multiplicative error bound. Nevertheless, because all the data points share a global residual covariance matrix, empirical experiments show that $\Sigma^{1/2}$ converges well on all the datasets.

Here we show that the variance of our sampling process is strictly lower than the rejection sampling.

**Theorem 3** *Here we follow the notation of equation(7) in the main paper. Let $\theta_1$ be the reject sampling estimator of $\Phi(0; -\mu, \Sigma)$, where $E[\theta_1] = E_{r \sim N(0, \Sigma)}[I\{r \preccurlyeq \mu\}]$. Let $\theta_2$ be the estimator of DMVP's sampling process, where $E[\theta_2] = E_{w \sim N(0, \Sigma_r)}[\Pr(z \preccurlyeq (w + \mu)|w)]$ and $z \sim N(0, V)$. We have $Var[\theta_2] < Var[\theta_1]$.*
*Proof.*

$$Var[\theta_2] = E[(\theta_2 - E[\theta_2])^2]$$
$$= E_{w \sim N(0, \Sigma_r)}[(\Pr(z \preccurlyeq (w + \mu)|w) - E[\theta_2])^2]$$
$$= E_{w \sim N(0, \Sigma_r)}[(E_{z \sim N(0, V)}[I\{z \preccurlyeq (w + \mu)\} - E[\theta_2]|w])^2]$$
$$< E_{w \sim N(0, \Sigma_r)}[E_{z \sim N(0, V)}[(I\{z \preccurlyeq (w + \mu)\} - E[\theta_2])^2|w]]$$
$$= E_{r \sim N(0, \Sigma)}[(I\{r \preccurlyeq \mu\} - E[\theta_1])^2]$$
$$\qquad (\textit{Here } r = z - w \textit{ and } E[\theta_1] = E[\theta_2])$$
$$= E[(\theta_1 - E[\theta_1])^2] = Var[\theta_1]$$

*The inequality follows the fact that $E[x^2] > E[x]^2$ given $Var[x] \neq 0$.*